\documentclass[journal,letterpaper]{IEEEtran}
\usepackage{lipsum} 

\usepackage{float}
\usepackage{amsmath,amsfonts,amssymb}
\usepackage{algorithmic}
\usepackage{array}
\usepackage[caption=false,font=normalsize,labelfont=sf,textfont=sf]{subfig}
\usepackage{pdfpages}
\usepackage{textcomp}
\usepackage{placeins}
\usepackage{stfloats}
\usepackage{url}
\usepackage{verbatim}
\usepackage{graphicx}
\usepackage{hyperref}
\usepackage{orcidlink}
\usepackage{listings}

\title{Simultaneous Localization and 3D-Semi Dense Mapping for Micro Drones Using Monocular Camera and Inertial Sensors}

  \author{
  \IEEEauthorblockN{
  Jeryes Danial ~\orcidlink{0000-0001-5428-3294}\textsuperscript{1},
  Yosi Ben Asher\textsuperscript{2},
  Itzik Klein \IEEEmembership{Senior Member,~IEEE}\IEEEauthorrefmark{1}~\orcidlink{0000-0001-7846-0654}\textsuperscript{1}
  }\\
  \IEEEauthorblockA{
  \textsuperscript{1}The Hatter Department of Marine Technologies, University of Haifa, 3498838, Israel
  }\\
  \IEEEauthorblockA{
  \textsuperscript{2}Computer Science, University of Haifa, 3498838, Israel
  }
  }
\date{}
\begin{document}

\maketitle

\begin{abstract}
	Monocular simultaneous localization and mapping (SLAM) algorithms estimate drone poses and build a 3D map using a single camera. Current algorithms include sparse methods that lack detailed geometry, while learning-driven approaches produce dense maps but are computationally intensive. Monocular SLAM also faces scale ambiguities, which affect its accuracy. To address these challenges, we propose an edge-aware lightweight monocular SLAM system combining sparse keypoint-based pose estimation with dense edge reconstruction. Our method employs deep learning-based depth prediction and edge detection, followed by optimization to refine keypoints and edges for geometric consistency, without relying on global loop closure or heavy neural computations. We fuse inertial data with vision by using an extended Kalman filter to resolve scale ambiguity and improve accuracy. The system operates in real time on low-power platforms, as demonstrated on a DJI Tello drone with a monocular camera and inertial sensors. In addition, we demonstrate robust autonomous navigation and obstacle avoidance in indoor corridors and on the TUM RGBD dataset. Our approach offers an effective, practical solution to real-time mapping and navigation in resource-constrained environments.
\end{abstract}

\section{Introduction}
Autonomous navigation in complex environments is a core challenge in robotics, particularly for resource-constrained aerial platforms such as micro-drones. A critical requirement for such systems is the ability to perform real-time, accurate 3D mapping and perception for navigation, obstacle avoidance, and interaction with the physical environment.

Classical visual SLAM systems, such as ORB-SLAM~\cite{mur2015orb,muglikar2020voxel}, rely on sparse feature matching, geometric triangulation, and global optimization to estimate camera pose and map structure. While effective for pose estimation, these methods often produce sparse maps "3D point" that lack structural richness, making them less suitable for tasks requiring dense 3D understanding or interaction,  While Voxel map ~\cite{muglikar2020voxel} offers improved retrieval and reasoning over visibility than ORB-SLAM ~\cite{mur2015orb}, it remains fundamentally sparse storing only a few landmark points per voxel and still lacks the structural and geometric richness required for dense perception and navigation. Moreover, their reliance on computationally intensive operations like loop closure ~\cite{mur2015orb,ho2007detecting} (a process that detects when the camera revisits a scene to improve accuracy) and global bundle adjustment \cite{triggs1999bundle} (an optimization technique that refines camera poses and 3D points simultaneously) can make real-time deployment on embedded systems impractical. Furthermore, Structure from Motion (SfM) is a foundational technique that reconstructs 3D structures from multiple images, serving as a precursor to SLAM. SfM begins by detecting and extracting features using methods such as SIFT \cite{lowe2004distinctive}, SURF \cite{bay2006surf}, ORB \cite{rublee2011orb}, or learned features \cite{agarwal2010bundle}. These features are matched across images, with outliers removed through geometric verification techniques like RANSAC \cite{hartley2003multiple}. Camera positions are estimated from these correspondences, and 3D points are triangulated to build a sparse reconstruction. Bundle adjustment then refines the entire structure by minimizing reprojection errors \cite{triggs1999bundle}. Despite its effectiveness, SfM faces similar challenges to SLAM in real-time applications, especially on resource-limited platforms, while Edge SLAM \cite{maity2017edge} has been proposed as an alternative that aims to overcome these limitations by generating semi-dense maps  but remains computationally heavy due to reliance on global optimization techniques such as loop closure. Additionally, their tracking loss function based on optical flow introduces inherent noise, which can degrade accuracy and hinder real-time performance on limited hardware.


Efficient visual-inertial odometry methods such as ~\cite{solodar2024vio,jung2024msckf,qin2018vins,ramezani2017omnidirectional}  employ EKF-based filtering techniques, which enable fast and efficient pose estimation and are commonly integrated into systems like ORB-SLAM to improve real-time performance. However, these methods typically generate sparse 3D point cloud maps, which can be insufficient for perception and navigation in complex environments.
Methods such as DeepTAM~\cite{zhou2018deeptam} generates dense depth maps and camera pose estimates using a deep neural network backbone with a large number of parameters. Yet, its limited accuracy, coupled with high floating-point operation counts and substantial model complexity, impairs real-time performance on resource-constrained devices lacking hardware acceleration.
While D3VO ~\cite{yang2020d3vo} achieves higher accuracy in depth and pose estimation by integrating deep learning with traditional visual odometry techniques, but this comes at the expense of increased computational complexity and large model parameters and weights, making it less suitable for real-time applications on less powerful devices. To address high-fidelity visual details and semantics learning-based approaches such as Radiance fields CNN "NeRFs"~\cite{mildenhall2021nerf} have shown impressive capabilities in dense 3D scene  reconstruction from monocular images. However, these models typically require high-end GPUs, assume static scenes, and lack real-time performance, especially on embedded hardware. Approaches like NeuralRecon~\cite{sun2021neuralrecon} and NICE-SLAM~\cite{zhu2022nice} fuse depth and pose estimation within neural architectures but are not computationally feasible for low-cost aerial systems.

To address the above mentioned monocular SLAM shortcomings, we propose a lightweight hybrid edge-aware monocular SLAM system designed for real-time operation on resource constrained platforms. Our method fuses classical sparse 3D keypoint triangulation with per-frame dense edge-point estimation from a lightweight neural network, specifically utilizing a pretrained FastDepth model based on the MobileNet-NNConv5 architecture with depth-wise separable layers  ~\cite{icra_2019_fastdepth}. These edge points are jointly optimized within a novel multi-loss bundle adjustment formulation. Further, to overcome the inherent scale ambiguity in monocular SLAM, we fuse the relative transformations estimated between image pairs obtained via epipolar geometry from consecutive frames with inertial measurements and the bundle adjustment pose estimation result. This fusion is performed using an extended Kalman filter (EKF), where the state vector represents pose estimation with respect to a global reference frame. The observations input to the EKF are relative transformations derived from visual data, which are inherently scale-ambiguous, while the API provides metric scale information and orientation cues. By integrating these sources, our approach effectively constrains the scale and improves pose accuracy over time. This ensures robust and consistent mapping suitable for real-time navigation on resource-constrained platforms.

The contributions of this paper are:
\begin{enumerate}
	\item A lightweight SLAM pipeline that integrates sparse epipolar geometry  with dense depth prediction and edge extraction for edge-anchored mapping.
	\item An edge-cycle consistency loss and a shape-aware structural constraint  that maintain coherence across views without requiring explicit 2D-2D edge matching.
	\item A multi-objective optimization (bundle adjustment) process that jointly refines camera poses, keypoints, and edge segments under local geometric constraints, eliminating the need for global loop closure or dense volumetric fusion.
\end{enumerate}

Our system provides real-time 3D mapping and accurate pose estimation on resource-limited platforms. By combining sparse keypoints with dense edge-aware depth, it enhances map detail and robustness, offering an efficient solution for perception and navigation in complex, cluttered environments suitable for embedded autonomous systems.

The remainder of this paper is organized as follows: Section ~\ref{sec:problem} presents the problem formulation. Section ~\ref{sec:approach} describes details  our proposed approach while  Section~\ref{sec:results} provides experimental validation and quantitative evaluation of our method. Finally, Section~\ref{sec:Conclusion} concludes the paper and outlines future research directions.

\section{Problem Formulation}
\label{sec:problem}
\noindent Traditional monocular SLAM systems, such as ORB-SLAM, typically follow a multi-stage pipeline involving 1) keypoint extraction and matching, where ORB keypoints and descriptors~\cite{rublee2011orb} are extracted from image frames and matched across temporally adjacent views, 2) motion estimation, where given matched keypoints points  $\mathbf{u}_i$ and $\mathbf{u}_j$ in consecutive frames for a single scene, the relative pose $\mathbf{T}_{i,j} \in$ SE(3) is computed using epipolar geometry~\cite{zhang1998determining}, with SE(3) representing the Special Euclidean Group in 3D that encodes rotations $R_{ij} \in$ SO(3) and translations vector $\mathbf{t}_{ij} \in \mathbb{R}^3$ into a single 4x4 matrix, 3) triangulation, where, given two or more calibrated views and the relative pose, 3D points $\mathbf{P}_i$ are triangulated via linear or iterative methods~\cite{zhang1998determining}, 4) local bundle adjustment, which jointly optimizes all camera poses $\{\mathbf{T}_{i,j}\}$ and map points $\mathbf{P}_i$ to minimize reprojection errors over all observations within a window of frames ~\cite{triggs1999bundle}, and 5) global optimization, such as loop closure, which is performed when the system revisits a previously mapped area to correct drift errors, typically using bundle adjustment to minimize the overall reprojection error.

Let  $\pi : \mathbb{R}^3 \rightarrow \mathbb{R}^2$ be the projection function that defined as:
\begin{equation}
\mathbf{u} = \pi\!\left(K \, \mathbf{P}\right), 
\label{eq:projection}
\end{equation}

where $\pi(\cdot)$ is the perspective projection function, $\mathbf{P} \in \mathbb{R}^3$ is a 3D point in the camera coordinate frame , $K \in \mathbb{R}^{3\times 3}$ denote the known intrinsic calibration (focal length and principle points) matrix of the camera, and $\mathbf{u} \in \mathbb{R}^2$ is the observation point in pixel unit. And let define an $N$ observation given set of 2d-2d correspondences $\{(\mathbf{u}_i^k, \mathbf{u}_j^k)\}_{k=1}^N$ between two frames, the relative pose estimation problem seeks $(R_{ij}, \mathbf{t}_{ij})$ that minimize the least-squares alignment error between the transformed points in frame $i$ and their corresponding points in frame $j$:
\begin{equation}
\mathcal{J}(R_{ij}, \mathbf{t}_{ij}) =
\sum_{k=1}^N \left\| R_{ij} \, \mathbf{u}_i^k + \mathbf{t}_{ij} - \mathbf{u}_j^k \right\|^2 .
\label{eq:relative_pose_cost}
\end{equation}
The optimal relative pose is then
\begin{equation}
(R_{ij}^\ast, \mathbf{t}_{ij}^\ast) = \arg\min_{R_{ij},\, \mathbf{t}_{ij}} \; \mathcal{J}(R_{ij}, \mathbf{t}_{ij}).
\label{eq:relative_pose_solution}
\end{equation}

Since the epipolar constraints are inherent in the geometric relationship between two sequence frames, the relative pose \((R_{ij}^\ast, \mathbf{t}_{ij}^\ast)\) can be estimated by decomposing the essential matrix  ~\cite{zhang1998determining}. These constraints provide a set of equations derived from point correspondences \((\mathbf{u}_i^k, \mathbf{u}_j^k)\), which can be expressed as a linear homogeneous system:

\begin{equation}
\mathbf{u}_j^{k\top} \, \mathbf{E}_{ij} \, \mathbf{u}_i^k = 0,
\label{eq:essentialM}
\end{equation}

where the essential matrix \(\mathbf{E}_{ij}\) encodes the intrinsic geometric relationship between the two frames and all the corresponding points adhere to these inherent constraints.
When camera intrinsics \(K\) are known, \(\mathbf{E}_{ij}\) can be decomposed into \(R_{ij}\) and \(\mathbf{t}_{ij}\). If intrinsics are unknown, we estimate the fundamental matrix \(\mathbf{F}_{ij}\) and recover \begin{equation}
\mathbf{E}_{ij} = K^\top \mathbf{F}_{ij} K\
\label{eq:fundemntalM}
\end{equation}. Solving this linear system and enforcing the constraints allows us to extract the relative pose.

With $(R_{ij}^\ast, \mathbf{t}_{ij}^\ast)$ estimated, the 3D coordinates of each matched feature can be recovered by triangulation. For a given correspondence $(\mathbf{u}_i^k, \mathbf{u}_j^k)$, we seek a 3D point $\mathbf{P}^k$ whose projections into both frames match the observed normalized coordinates:

\begin{multline}
\mathbf{P}^{k\ast} = \arg\min_{\mathbf{P}^k} \left\| \mathbf{u}_i^k - \pi\!\left( K \, \mathbf{P}^k \right) \right\|^2 \\
+ \left\| \mathbf{u}_j^k - \pi\!\left( K \left[ R_{ij}^\ast \mathbf{P}^k + \mathbf{t}_{ij}^\ast \right] \right) \right\|^2
\label{eq:triangulation}
\end{multline}

Once an initial set of camera poses $\{\mathbf{T}_i\}$ and 3D points $\{\mathbf{P}^k\}$ is obtained, we refine them jointly via local bundle adjustment within a window of frames. The objective function is the total reprojection error over all visible points:
\begin{equation}
\mathcal{L}_{\text{reproj}} =
\sum_{i} \sum_{k \in \mathcal{V}(i)}
\left\| \mathbf{u}_{ik} - \pi\!\left( K \, \mathbf{T}_i \, \mathbf{P}^k \right) \right\|^2 ,
\label{eq:bundle_adjustment}
\end{equation}
where $\mathcal{V}(i)$ denotes the set of points visible in frame $i$. This formulation ensures that the estimated poses and 3D structure are consistent with all available observations, forming the geometric foundation of our SLAM system.
Finally, to mitigate accumulated drift, a global bundle adjustment is performed over the entire looped trajectory once a loop closure is detected. This optimization refines all camera poses and the sparse 3D map simultaneously, leveraging multiple frames to enforce global consistency. The process involves minimizing a reprojection error similar to~\eqref{eq:bundle_adjustment}, but extended across the entire looped sequence to ensure accurate and drift-free reconstruction.

It is well known that even classical sparse-feature SLAM systems such as ORB-SLAM may suffer from degeneracy when inter-frame motion is small, parallax is insufficient, or the majority of points lie on a single planar surface. In such cases, the essential matrix estimation in~\eqref{eq:fundemntalM} and triangulation in~\eqref{eq:triangulation} become ill-conditioned, yet global bundle adjustment after loop closure can often recover accurate poses and structure.

In our setting, the challenge is increased because we are extracting a semi-dense edge map instead of sparse keypoints. Since the number of features is much larger than the number of keypoints, each stage of the processing pipeline becomes significantly more computationally demanding. For instance, searching for correspondences between images scales quadratically with the number of features, and fitting the essential matrix with RANSAC requires more iterations due to the higher number of outliers. Additionally, triangulation involves solving a larger linear system. During global bundle adjustment, the number of parameters to optimize increases substantially, which causes the optimization process to take longer. Overall, these factors make it difficult to achieve real-time performance on lightweight platforms. The main challenge is to develop a method that can provide accurate dense mapping and robust pose estimation despite the computational load introduced by large sets of edge-based correspondences.

\section{Proposed Approach}
\begin{figure}[!htbp]
	\centering
	\includegraphics[width=\linewidth]{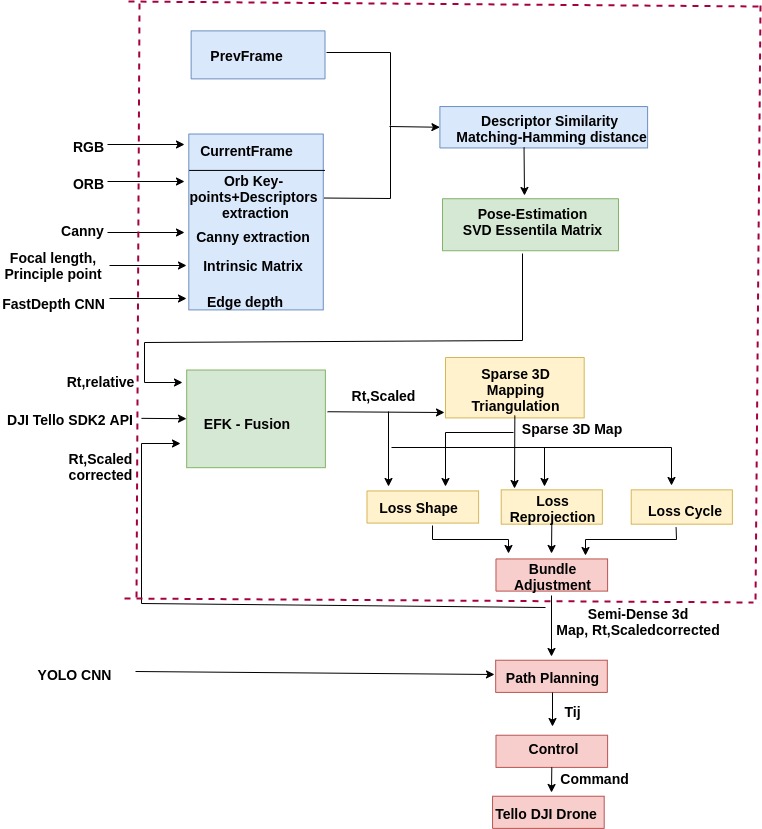}
	\caption{\label{fig:edgeAwareSLAM} Pipeline for our proposed edge-aware constraint SLAM approach.}
\end{figure}

\noindent Our approach aims to facilitate real-time, semi-dense 3D mapping on resource-constrained aerial platforms by integrating standard geometric vision with lightweight deep learning modules. Instead of relying on global optimization techniques such as loop closure and global bundle adjustment, or on computationally intensive deep learning methods like D3VO that are unsuitable for limited hardware, we develop a local geometry-aware optimization strategy. This strategy leverages edge-aware constraints and learned monocular depth to achieve efficient mapping.
Figure ~\ref{fig:edgeAwareSLAM} highlights four different threads in our proposed pipeline: 1. Image processing and feature extraction (in blue), 2. Pose estimation and sensor fusion (in green), 3. Semi-dense edge map and 3D anchors generation (in yellow), and 4) Edge-aware local optimization (in pink). In the following subsections we elaborate on each thread.

\label{sec:approach}

\subsection{Preprocessing}

The preprocessing and sensor fusion part contains two Threads (1 and 2) as described below.
\noindent \textbf{Thread 1 — Image Pre-Processing and Feature Extraction:} This thread handles the preprocessing of incoming images and the extraction of key visual information. ORB keypoints and Canny edges are detected in each frame. The ORB descriptors are then matched across frames using Hamming distance~\cite{lim2014real}, which efficiently measures the similarity between binary feature descriptors. To accelerate this process, the descriptors are stored in a KD-tree~\cite{gehrig2017visual}, a data structure optimized for fast nearest-neighbor searches, crucial for real-time performance. Concurrently, a lightweight FastDepth CNN~\cite{icra_2019_fastdepth} predicts per-frame dense depth maps, with edge pixels associated with their corresponding depth values. Camera intrinsics are also incorporated as input during this stage.

\textbf{Thread 2 — Pose Estimation and Sensor Fusion:} To estimate accurate and scaled camera poses, this thread fuses the  relative motion estimation ~\ref{sec:problem} $R_{ij} \in SO(3)$ and $\mathbf{t}_{ij} \in \mathbb{R}^3$  with  inertial measurements using an extended Kalman filter (EKF) ~\cite{enwiki:1299274411}. The relative motion is computed by estimating the essential matrix between frame pairs via RANSAC~\cite{hartley2003multiple} and SVD decomposition~\cite{zhang1998determining}, based on the epipolar geometry derived from the matched ORB features. The essential matrix \textbf{E} ~\eqref{eq:essentialM}, describes the geometric relationship between two camera views, specifically how they are rotated and shifted relative to each other. In cases where the camera is uncalibrated, the fundamental matrix \textbf{F} can be used ~\eqref{eq:fundemntalM}, which relates the two views without requiring known intrinsics ~\cite{zhang1998determining}. RANSAC is used to robustly identify inlier feature matches by filtering out outliers, while SVD decomposition is employed to accurately recover the relative rotation and translation from the essential matrix, both of which are crucial for reliable pose estimation in the presence of noise and outliers.

The EKF fuses the monocular relative pose estimates  with the linear velocity and Euler angles provided by the constrained resources drone API (DJI Tello SDK2 API ~\cite{giernacki2022dji}). The system operates under the constraint that the data from the SDK is more frequent but subject to drift, while the monocular relative pose is sparse but accurate over time.

For the fusion process the following state vector is defined: \begin{equation}
\mathbf{x} = 
\begin{bmatrix}
\boldsymbol{\scriptstyle p} \\
\boldsymbol{\alpha}
\label{eq:stateV}
\end{bmatrix}
=
\begin{bmatrix}
x & y & z & \phi & \theta & \psi
\end{bmatrix}^T
\end{equation},where $\mathbf{\boldsymbol{\scriptstyle p}} \in \mathbb{R}^3$ is the global position  $\mathbf{x,y,z}$ , and $\boldsymbol{\alpha} = (\phi, \theta, \psi)$ are the Euler angles representing roll, pitch, and yaw. The drone's linear velocity from the API  propagate the position: \begin{equation}
\mathbf{\boldsymbol{\scriptstyle p}}_{k|k-1} = \mathbf{\boldsymbol{\scriptstyle p}}_{k-1} + \mathbf{R}_{\text{imu}}(\boldsymbol{\alpha}_{k-1}) \cdot \mathbf{v}_{\text{imu}} \cdot \Delta t
\end{equation},where $\mathbf{v}_{\text{imu}} \in \mathbb{R}^3$ is the velocity vector expressed in the body frame,$\mathbf{R}_{\text{imu}}$ is the rotation matrix from the body/IMU frame to the world frame-constructed from the Euler angles $\boldsymbol{\alpha}_{k-1}$, and $\mathbf{\Delta t}$ is the time interval between the two state updates.

We modeled the process noise covariance based on the noise characteristics of the SDK2 API data, since it is not fully accurate and drifts over time especially as the battery level decreases. The motivation to do so came from analyzing our laboratory experiments. When the drone followed a simple straight line trajectory without changing orientation, the API’s estimates became less reliable as the battery drained. These validations showed that API data accuracy decreased over time and with lower battery levels. Therefore, we define the process noise covariance as a function of the known battery level $b_k \in [0, 1]$  (via the SDK2 API) and the time $\tau$ since the last monocular update: \begin{equation}
\mathbf{Q}_k = \beta \cdot (1 - b_k + \lambda \tau) \cdot \mathbf{I}_6
\end{equation}
where $\beta$ and $\lambda$ are tuning constants that scale the noise based on energy degradation and time  $\mathbf{\tau}$ since last correction. The process noise covariance $\mathbf{Q}_k \in \mathbb{R}^{6\times6}$ models uncertainty in both position and orientation. In the update step, the first measurement involves the Euler angles obtained directly from the SDK: \begin{equation}
\mathbf{z}_\text{api} = \boldsymbol{\alpha}_{\text{api}} \in \mathbb{R}^3
\end{equation}
where $\mathbf{z}_\text{api}$ is the measurement vector containing the roll, pitch, and yaw angles obtained from the SDK.

The measurement matrix is:

\begin{equation}
\mathbf{H}_{\text{api}} = 
\begin{bmatrix}
\mathbf{0}_{3 \times 3} & \mathbf{I}_3
\end{bmatrix}
\label{eq:Hapi}
\end{equation}
where $\mathbf{0}_{3 \times 3}$ is a $3 \times 3$ zero matrix indicating no direct dependence of the measured angles on the first three state variables, and $\mathbf{I}_3$ is the $3 \times 3$ identity matrix, representing the direct relationship between the measured Euler angles and the corresponding components of the state vector.

The second measurement comes from monocular visual odometry, which provides the relative pose estimation $\Delta \mathbf{T}_{\text{rel}} = [\Delta \mathbf{R}, \Delta \mathbf{t}]$ between two frames. Here, $\Delta \mathbf{T}_{\text{rel}} \in$ SE(3) represents the relative transformation, with $\Delta \mathbf{R} \in$ SO(3) being the relative rotation matrix and $\Delta \mathbf{t} \in \mathbb{R}^3$ the relative translation vector.

To estimate the global pose, we accumulate the relative poses:

\begin{equation}
\mathbf{X}_k = \mathbf{X}_{k-1} \cdot \Delta \mathbf{T}_{\text{rel}}
\end{equation}

We then define the measurement vector:

\begin{equation}
\mathbf{z}_{\text{vo}} = 
\begin{bmatrix}
\mathbf{\boldsymbol{\scriptstyle p}}_{\text{vo}} \\
\boldsymbol{\alpha}_{\text{vo}}
\end{bmatrix}
\in \mathbb{R}^6
\label{eq:OBSapi}
\end{equation}, and we combine \eqref{eq:Hapi} and \eqref{eq:OBSapi} to a full-state measurement.

\subsection{Geometry-Aware Optimization}

The geometry-aware optimization includes Thread 3 and 4 as described below.

\textbf{Thread 3 — Dense Edge Map and 3D Anchors Generation:} 
Using the known depth maps (from FastDepth CNN pre-trianed model ~\cite{icra_2019_fastdepth}) and the estimated camera poses ~\eqref{eq:stateV}, this thread reconstructs a set of 3D points (referred to as \emph{anchors}) by triangulating corresponding features. 

Loss functions are used here to enforce geometric and structural constraints, enabling accurate local reconstruction and preventing global drift. We define three local losses: (1) sparse reprojection error from triangulated points $L_{\mathrm{reproj}}$, (2) cycle consistency loss from multi-view edge projections $L_{\mathrm{cycle}}$, and (3) shape-preserving loss for structural edge coherence $L_{\mathrm{shape}}$.

The \textbf{reprojection loss} defined as :
\begin{equation}
\mathcal{L}_{\text{reproj}} = \sum_{i,k} \left\| \mathbf{u}_{ik} - \mathbf{u}_{ik}^{\text{proj}} \right\|^2,
\label{eq:Lrep}
\end{equation}
where \( \mathbf{u}_{ik} \) is the observed pixel location of the 3D scene point \( P_k \) in frame \( i \), and \( \mathbf{u}_{ik}^{\text{proj}} \) is the projected location of \( P_k \) in frame \( i \).  

The projected location is given by:

\begin{equation}
\mathbf{u}_{ik}^{\text{proj}} = \pi ( R_i P_k + t_i ),
\end{equation}
where \( R_i \) is the rotation matrix for frame \( i \) and \( t_i \) is the translation vector for frame \( i \) and \( \pi \) is a projection function that map a 3D point into 2D pixel point ~\eqref{eq:projection}.
This loss directly measures the residual between observed image points and projected 3D points, enforcing geometric consistency. Minimizing it aligns the estimated scene structure and camera poses with actual observations, improving accuracy. Mathematically, it creates a differentiable residual that guides parameter optimization toward the true scene geometry.

The \textbf{cycle-consistency loss}  for edge pixels uses depth and relative pose to create a closed transformation cycle:

\begin{equation}
\mathbf{P}_i = \pi^{-1}(\mathbf{u^*}_i, d_i) = d_i \cdot \mathbf{K}^{-1} \cdot \mathbf{u^*}_i,
\end{equation}
where \( \mathbf{u^*}_i \in \mathbb{R}^2 \) is a 2D pixel coordinate of an edge pixel in frame \( i \), and \( d_i \) is the depth at that pixel, with 
\( \mathbf{K}^{-1} \) being the inverse camera intrinsic matrix and \( \pi^{-1} \) the back-projection function.

Subsequently-, the 3D point is transformed to frame \( j \) as:

\begin{equation}
\mathbf{P}_j = \mathbf{T}_{i,j} \cdot \mathbf{P}_i,
\end{equation}
where \( \mathbf{T}_{i,j} \in\)  SE(3)  is a transformation matrix from frame \( i \) to \( j \), \( \mathbf{P}_i \) is the  3D point in frame \( i \), and \( \mathbf{P}_j \) is a 3D point in frame \( j \). This 3D point in frame \( j \) is then projected into the image space via:

\begin{equation}
\mathbf{u}_j = \pi(\mathbf{P}_j),
\end{equation}
where \( \pi \) is the projection function , \( \mathbf{u}_j \) is a  projected pixel in frame \( j \).

Next, the projected pixel \( \mathbf{u}_j \) is back-projected into 3D space using the depth map \( d_j \):

\begin{equation}
P'_{j} = \pi^{-1}(\mathbf{u}_j, d_j),
\end{equation}
where \( d_j \) is the depth map in frame \( j \).  
\( P'_{j} \) is a back-projected 3D point from frame \( j \).

To complete a full cycle transformation , the 3D point  \( P'_{j} \) in frame \( j\) should be transformed back to frame  \( i \):

\begin{equation}
P'_{i} = \mathbf{T}_{i,j}^{-1} \cdot P'_{j},
\end{equation}

Finally, the last step is to reproject \( P'_{i} \) as a pixel into frame \(i\):

\begin{equation}
u'_{i} = \pi( P'_{i}),
\end{equation}
where \( u'_{i} \) is the reprojected pixel in frame \( i \).  

The full cycle-consistency loss :

\begin{equation}
\mathcal{L}_{\text{cycle}} = \sum_{\mathbf{u^*}_i \in \mathcal{E}} \left\| \mathbf{u^*}_i - \mathbf{u}_i' \right\|^2,
\label{eq:Lcyc}
\end{equation}
where \( \mathbf{u^*}_i \) is the observed edge pixel and  \( \mathcal{E} \) is a  set of edge pixels. We use this loss to enforce a strong geometric constraint that the round-trip transformation preserves pixel locations, regularizing camera pose and depth consistency. Mathematically, minimizing this residual ensures the composition of transformations approaches the identity, leading to more accurate scene reconstructions.

\begin{figure}[!htbp]
	\centering
	\begin{minipage}[b]{0.22\textwidth}
		\centering
		\includegraphics[width=\linewidth]{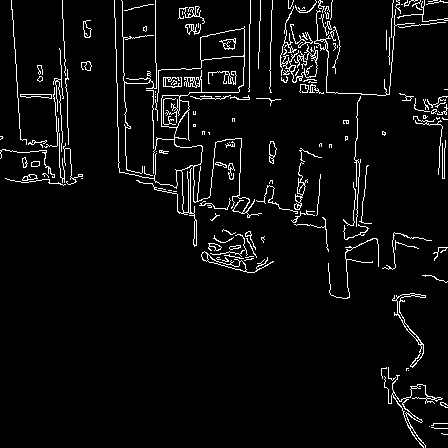}
	\end{minipage}
	\hfill
	\begin{minipage}[b]{0.22\textwidth}
		\centering
		\includegraphics[width=\linewidth]{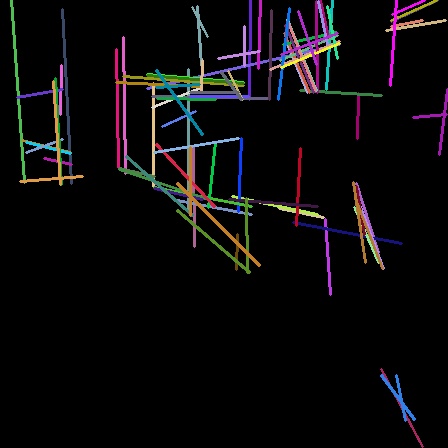}
	\end{minipage}
	\caption{Semantic constraint in the L-shape, where intersecting lines detected by  Canny edges detection (left) and a pair of and reprojected lines that intersect each other, forming approximate L shapes (right).}
	\label{fig:intersection}
\end{figure}

To encourage structural regularity in indoor environments, we introduce a geometric \textbf{L-shape structural} loss  based on edge-based L-shape constraints. to evaluate the \begin{enumerate}
	\item An  angle consistency at edge corners \textbf{(Langle)} 
	\item Collinearity of edge pixels along straight edge segments \textbf{(Lcollinear)}
\end{enumerate}. For each detected corner from Canny edges, we identify two intersecting edge directions forming an approximate L-shape , using corresponding depth and pose information as illustrated in Figure ~\ref{fig:intersection}.

The angle loss defined by:

\begin{equation}
\mathcal{L}_{\text{angle}} = \frac{1}{N} \sum_{i=1}^{N} \left( 
\cos(\theta_{\text{proj}}^{(i)}) - \cos(\theta_{\text{expected}}^{(i)}) 
\right)^2,
\label{eq:Lang}
\end{equation}
where \( \theta_{\text{proj}}^{(i)} \) is the projected angle, \( \theta_{\text{expected}}^{(i)} \) is the expected angle from image edges,
\( N \) is the number of L-shape junctions, and the  projected angle is:

\begin{equation}
\mathcal{\theta}_{\text{proj}, i} =  \arccos \left( \frac{ \vec{v}_{1,i}^{(3D),\text{proj}} \cdot \vec{v}_{2,i}^{(3D),\text{proj}} }{ \|\vec{v}_{1,i}^{(3D),\text{proj}}\| \|\vec{v}_{2,i}^{(3D),\text{proj}}\| } \right),
\end{equation}
Here \( \vec{v}_{1,i}^{(3D),\text{proj}} \) and \( \vec{v}_{2,i}^{(3D),\text{proj}} \) projection vector forming an approximate L-shape after projection. 

The \textbf{Colinearity Loss} is defined by:

\begin{equation}
\mathcal{L}_{\text{colinear}} = \frac{1}{N} \sum_{i=1}^{N} \left[
\frac{1}{M_1^{(i)}} \sum_{j=1}^{M_1^{(i)}} d_{j,1}^2 +
\frac{1}{M_2^{(i)}} \sum_{k=1}^{M_2^{(i)}} d_{k,2}^2 
\right]
\label{eq:Lcol}
\end{equation}

\noindent where \( M_1^{(i)} \), and \( M_2^{(i)} \) are the number of sampled edge points on the two segments forming the \( i \)-th L-shape and  the \( d_{j,1} \), and \( d_{k,2} \) are the perpendicular distances from edge points to their best-fit 3D lines in each lines.
The total loss combines the angle and collinearity:

\begin{equation}
\mathcal{L}_{\text{Lshape}} = \lambda_{\theta} \mathcal{L}_{\text{angle}} + \lambda_{\text{col}} \mathcal{L}_{\text{colinear}},
\label{eq:Lshp}
\end{equation}
where \( \lambda_{\theta} \) is the weighting factor that controls the relative importance of the angle consistency term, and 
\( \lambda_{\text{col}} \) is the weighting factor that controls the influence of the colinearity regularization.
Finally, by combining~\eqref{eq:Lrep},~\eqref{eq:Lcyc}, and~\eqref{eq:Lshp}, the total edge-aware SLAM loss is defined by:
\begin{equation}
\label{equ:Ltotal}
\min_{\mathcal{P}_w, \mathcal{T}_w, \mathcal{D}_w} \mathcal{L}_{\text{total}} = \lambda_{\text{reproj}} \mathcal{L}_{\text{reproj}} + \lambda_{\text{cycle}} \mathcal{L}_{\text{cycle}} + \lambda_{\text{shape}} \mathcal{L}_{\text{Lshape}},
\end{equation}
where \(\lambda_{\text{reproj}}\), \(\lambda_{\text{cycle}}\), and \(\lambda_{\text{shape}}\) are scalar weights controlling the contribution of the reprojection, cycle, and shape losses, respectively.

\textbf{Thread 4 — Edge-Aware Local Optimization}

\noindent We implement a local edge-aware bundle adjustment (BA) framework that simultaneously estimates camera poses, 3D point locations, and depth edge maps within a short sliding temporal window, aiming to minimize the total loss as described in~\eqref{equ:Ltotal}. Minimizing this least squares error ensures that the estimated parameters best fit the observed data, leading to more accurate and reliable local geometric reconstructions. Unlike traditional SLAM systems that depend on loop closure and global optimization, our method focuses on strong local geometric consistency, enabling efficient and dense edge-map reconstruction in real time. The optimization problem is solved using the Levenberg–Marquardt (LM) algorithm ~\cite{ranganathan2004levenberg}, which balances Gauss–Newton (GN) ~\cite{wang2012gauss} and gradient descent (GD) ~\cite{ruder2016overview}. 
We want to minimize the nonlinear least squares ~\cite{clark2018learning} total loss as described in ~\eqref{equ:Ltotal}.
During the optimization process, several challenges can arise:

\noindent \begin{enumerate}
	\item \textbf{Nonlinearity:}  
	The cost function involves inherently nonlinear projection models and geometric constraints, which make the optimization landscape non-convex. This nonlinearity can cause the algorithm to become trapped in local minima, preventing convergence to the optimal solution.
	\item \textbf{Singularities:}  
	Singular configurations occur when certain geometric arrangements, such as collinear points or planar surfaces, cause the Jacobian matrix to lose rank. This leads to instability and unreliable parameter estimates, especially in environments with repetitive or symmetrical structures.
	\item \textbf{Rank deficiency:}  
	When the observed data lacks sufficient geometric diversity—such as regions with feature-sparse environments or minimal camera movement—the system becomes rank-deficient. This results in weak gradients and poor convergence, making accurate reconstruction difficult.
	\item \textbf{Imbalance in scale weights:}  
	Our multi-loss formulation employs different scale weights (\(\mathcal{\lambda}_{\text{reproj}}\), \(\mathcal{\lambda}_{\text{cycle}}\), \(\mathcal{\lambda}_{\text{shape}}\)) to balance their respective contributions. If these weights are not properly tuned, one loss term may dominate the others, leading to an unbalanced optimization process that hampers overall accuracy and convergence stability.
\end{enumerate}
Our system addresses key LS challenges as follows: \textbf{Nonlinearity} is stabilized using regularization and the Levenberg–Marquardt algorithm. \textbf{Singularities} are prevented by applying diagonal regularization (\(\mu I\)), ensuring invertibility. \textbf{Ill-conditioned Jacobians} are improved through oblique camera motions, like zigzag trajectories, to enhance parallax. \textbf{Loss imbalance} is managed with adaptive weighting based on uncertainty, balancing reprojection, cycle, and shape constraints for stable updates.

\section{Analysis and Results}
\label{sec:results}
\noindent In this section, we present experimental results to demonstrate the effectiveness of our proposed edge-aware SLAM system. 

\subsection*{A. Dataset}

The TUM RGB-D benchmark\cite{endres20133,sturm2012benchmark} provides synchronized RGB-D sequences in indoor environments. Derived from the collection by the TUM CVPR team and released under a Creative Commons license, it includes 23 sequences with diverse motions, viewpoints, and illumination. Ranging from 2 to 10 minutes, these sequences enable robust testing across short, fast, and long, complex paths, with ground-truth(GT) poses for accurate quantitative assessment.
For depth estimation, we utilize the FastDepth model trained on the NYU Depth v2 dataset \cite{silberman2012indoor}. FastDepth employs a lightweight encoder-decoder architecture with MobileNet \cite{deng2009imagenet} as the backbone, making it suitable for real-time applications. Its encoder uses depthwise separable convolutions to reduce complexity, and skip connections help preserve image details during decoding. The model’s compact design enables efficient inference on embedded hardware, such as the Jetson TX2.

\subsection*{B. Evaluation Metrics}

To quantitatively assess trajectory accuracy, two widely used metrics were adopted: the absolute pose error (APE) and the absolute trajectory error (ATE).

\paragraph{Absolute pose error:} measures the instantaneous Euclidean distance between the estimated and ground-truth camera positions at each timestamp:
\begin{equation}
\mathrm{APE}_i = \left\| \mathbf{t}_i^{\mathrm{est}} - \mathbf{t}_i^{\mathrm{gt}} \right\|_2
\label{eq:ape}
\end{equation}
where $\mathbf{t}_i^{\mathrm{est}} \in \mathbb{R}^3$ is the estimated position, $\mathbf{t}_i^{\mathrm{gt}} \in \mathbb{R}^3$ is the GT position, and $\| \cdot \|_2$ is the Euclidean norm.

\paragraph{Absolute trajectory error:} evaluates the global drift over the entire sequence, considering both translation and rotation:
\begin{equation}
\mathrm{ATE}_{\mathrm{RMS}} =
\sqrt{\frac{1}{N} \sum_{i=1}^{N}
	\left\| \mathbf{T}_i^{\mathrm{gt}^{-1}} \mathbf{T}_i^{\mathrm{est}} \right\|_F^2 }
\label{eq:ate}
\end{equation}
where $\mathbf{T}_i^{\mathrm{est}}, \mathbf{T}_i^{\mathrm{gt}} \in SE(3)$ are the estimated and GT camera poses at frame $i$, $N$ is the total number of frames, and $\| \cdot \|_F$ denotes the Frobenius norm. This metric captures accumulated translational and rotational deviations across the sequence.

\subsection*{C. Positioning Performance Analysis}

This section presents a quantitative evaluation of the trajectory accuracy achieved by our system on the TUM RGB-D dataset, comparing it with a baseline method.

\begin{table}[!htbp]
	\centering
	\caption{Quantitative evaluation of the ATE metric on the TUM RGB-D dataset. Lower values indicate higher accuracy.}
	\label{tab:results}
	\begin{tabular}{|l|c|c|c|}
		\hline
		\textbf{Method} & \textbf{RMSE [m]} & \textbf{Mean [m]} & \textbf{Std [m]} \\
		\hline
		ORB-SLAM2 (baseline) & 0.182 & 0.17 & 0.71 \\
		\textbf{Edge-Aware SLAM (ours)} & \textbf{0.046} & \textbf{0.040} & \textbf{0.011} \\
		\hline
		\textbf{Improvement} & \textbf{74.7\%} & \textbf{76.5\%} & \textbf{98.4\%}
		\\
		\hline
	\end{tabular}
\end{table}

\begin{figure*}[!htbp]
	\centering
	\begin{minipage}[b]{0.45\linewidth}
		\centering
		\includegraphics[width=\linewidth]{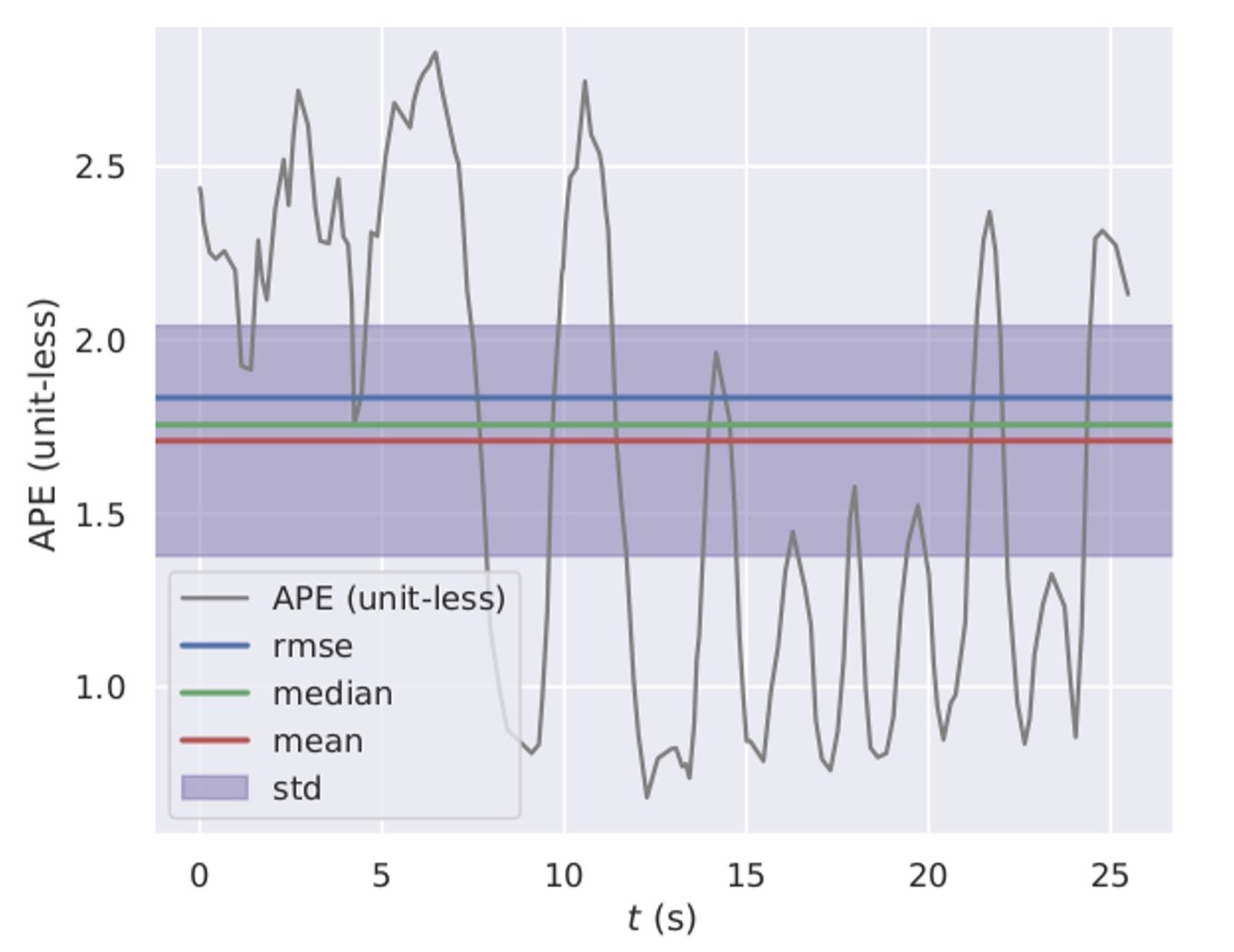}
		\label{fig:ate1}
	\end{minipage}
	\hspace{0.05\linewidth} 
	\begin{minipage}[b]{0.45\linewidth}
		\centering
		\includegraphics[width=\linewidth]{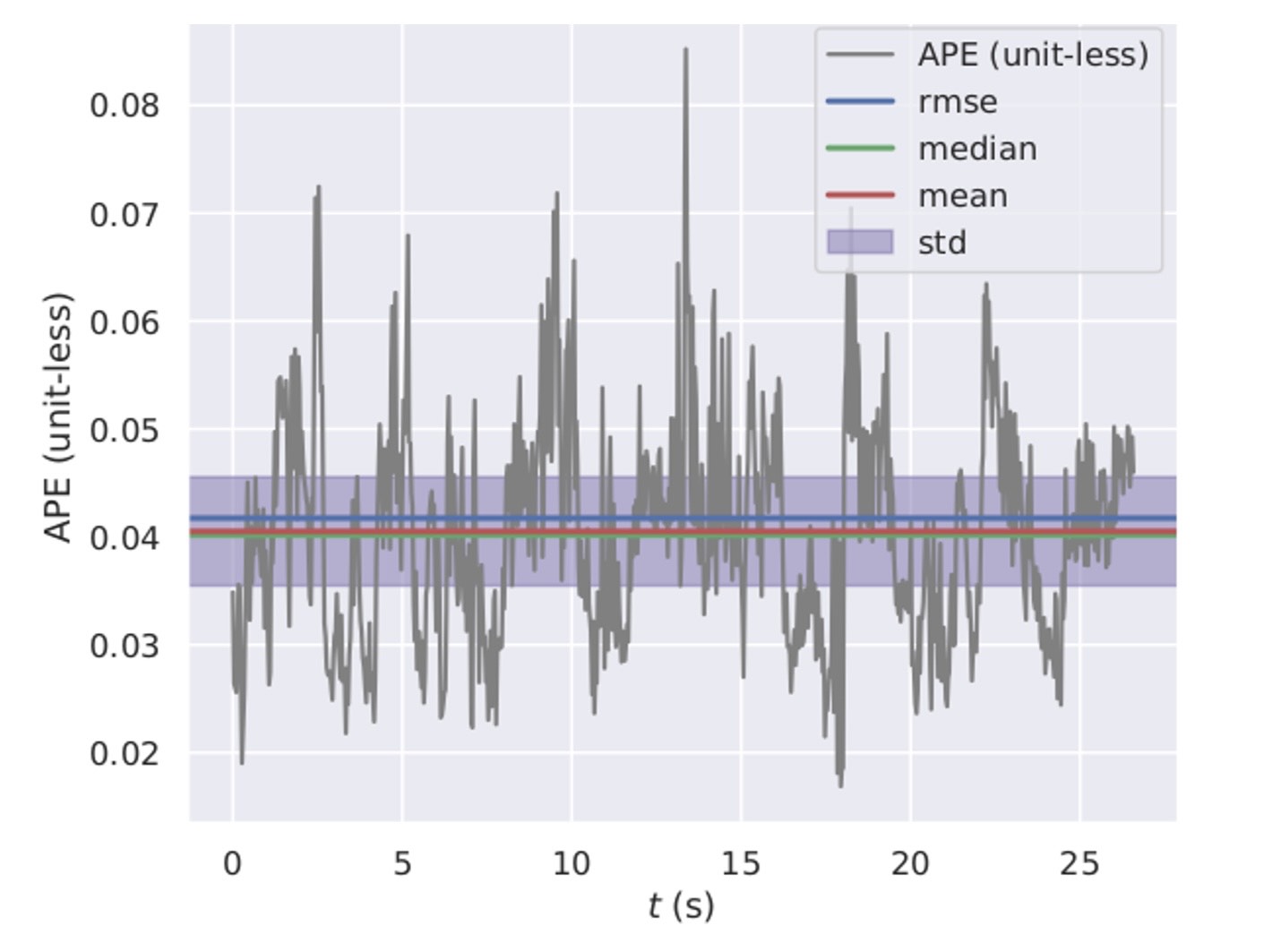}
		\label{fig:ate2}
	\end{minipage}
	\caption{APE w.r.t. full transformation (unitless), (left) Edge-Aware SLAM (ours) and (right) ORB-SLAM2 (baseline)}
	\label{fig:slam_comparison}
\end{figure*}

Table \ref{tab:results} presents a quantitative evolution of the ATE metric showing its root mean square error (RMSE), mean, and standard deviation (STD). The evaluation was made on the TUM RGB-D dataset. Results  show our system outperforms ORB-SLAM2 in ATE, with lower RMSE, mean, and std, indicating better accuracy and stability. Figure \ref{fig:slam_comparison} confirms this, showing our method maintains more stable pose estimates throughout the sequence.

\subsection*{D. Qualitative 3D Mapping Evaluation}

To assess the spatial accuracy and detail of the reconstructed maps, we perform visual inspections using tools such as Pangolin and ORB-SLAM2 frameworks. These visualizations enable an intuitive comparison of the map quality, including the level of detail, completeness, and correctness of the reconstructed environments.

Our system demonstrates robust performance across various challenging environments, successfully navigating complex areas, identifying pathways, and locating exits. These capabilities validate the practical accuracy and reliability of the generated maps in real-world scenarios.

\begin{figure}[!htbp]
	\centering
	\begin{minipage}[t]{0.24\textwidth}
		\centering
		\includegraphics[width=\linewidth]{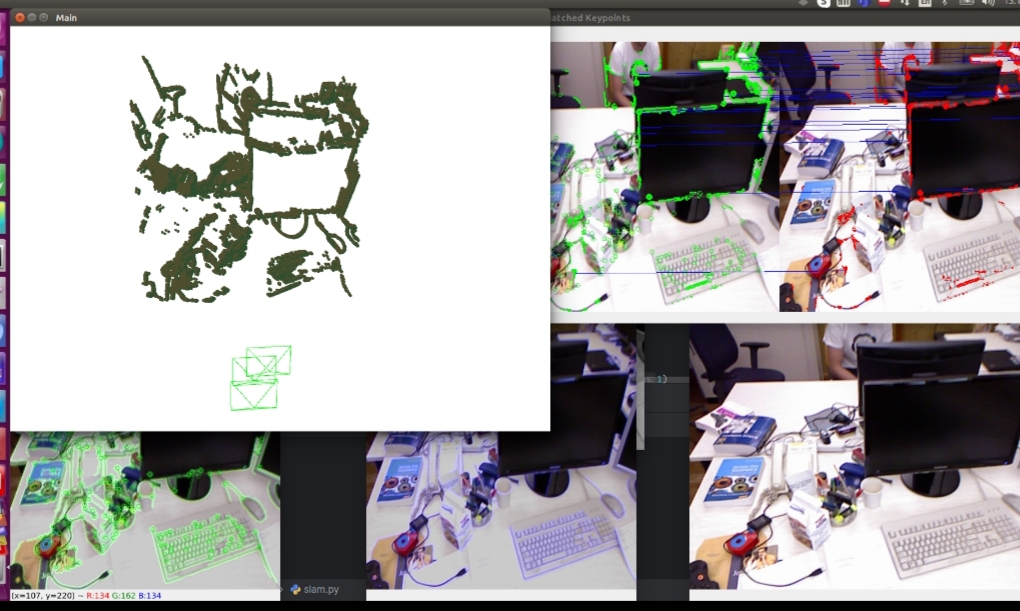}
		\label{fig:error_map}
	\end{minipage}
	\hfill
	\begin{minipage}[t]{0.24\textwidth}
		\centering
		\includegraphics[width=\linewidth]{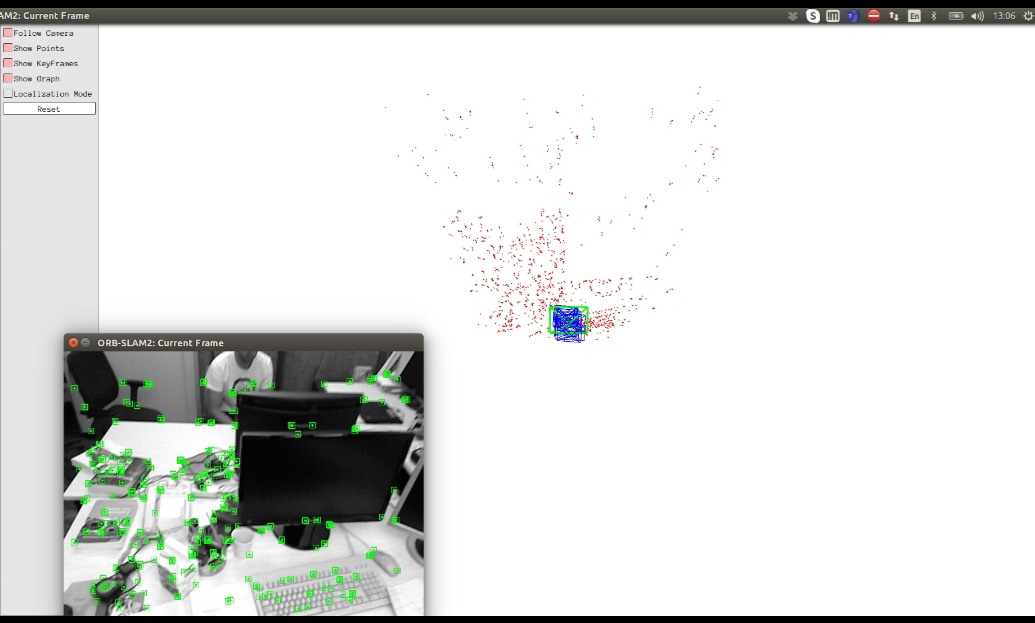}
		\label{fig:orbslam2_result}
	\end{minipage}
	\caption{(left) 3D map from our Edge-SLAM during initial frames in TUM and (right) 3D point cloud from ORB-SLAM2. Our method produces a clear, accurate 3D map early on, whereas ORB-SLAM2's point cloud is less interpretable at this stage.}
	\label{fig:initial_map}
\end{figure}

Figure \ref{fig:initial_map} illustrates the early-stage mapping performance, comparing our edge-aware SLAM’s 3D edge map with the reprojected edges of ORB-SLAM2. It highlights the accuracy and speed advantages of our method. Meanwhile, Fig.~\ref{fig:initial_map} showcases the initial mapping quality shortly after starting the system, demonstrating its ability to produce clear and detailed maps early on.

\begin{figure}[H]
	\centering
	\begin{minipage}[t]{0.24\textwidth}
		\centering
		\includegraphics[width=\linewidth]{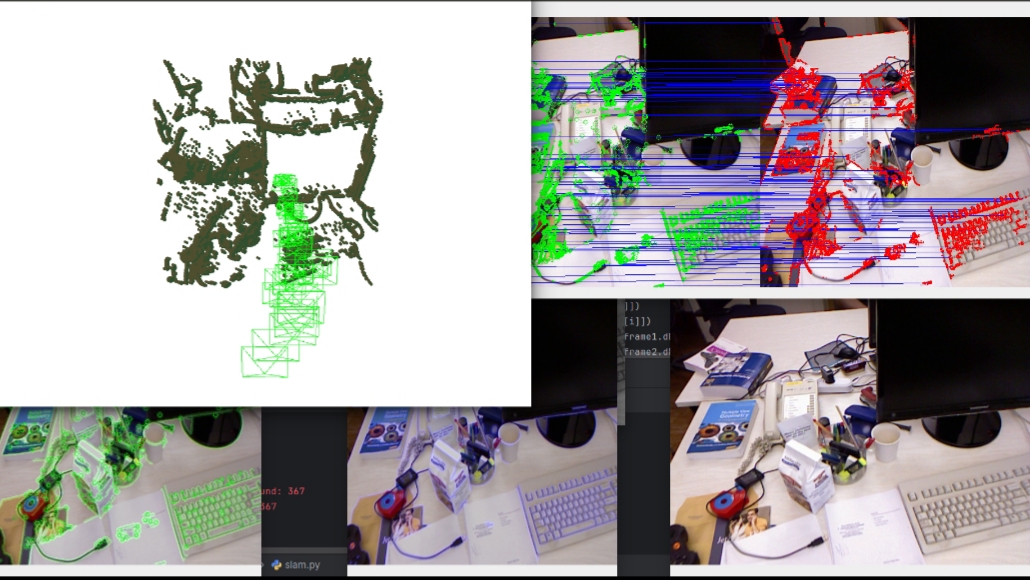}
		\label{fig:error_map}
	\end{minipage}
	\hfill
	\begin{minipage}[t]{0.24\textwidth}
		\centering
		\includegraphics[width=\linewidth]{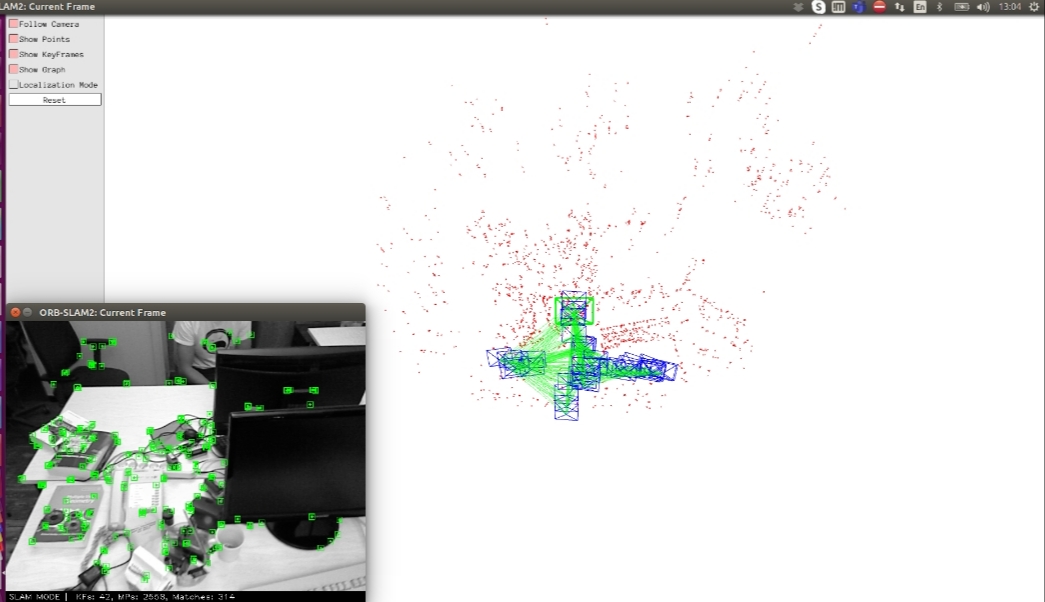}
		\label{fig:orbslam2_result}
	\end{minipage}
	\caption{(left) Our complete map after processing all frames, showing high accuracy and no drift and (right) ORB-SLAM2’s map, less clear and less interpretable. Our system produces a more accurate, coherent map over time, unlike ORB-SLAM2.}
	\label{fig:full_map_tum}
\end{figure}

\begin{figure}[H]
	\centering
	\begin{minipage}[b]{0.24\textwidth}
		\centering
		\includegraphics[width=\linewidth]{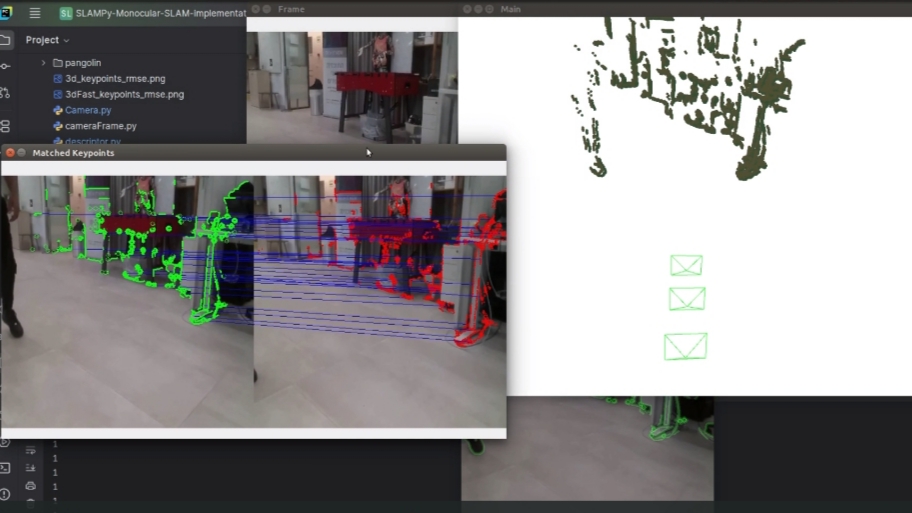}
		\label{fig:error_map}
	\end{minipage}
	\hfill
	\begin{minipage}[b]{0.24\textwidth}
		\centering
		\includegraphics[width=\linewidth]{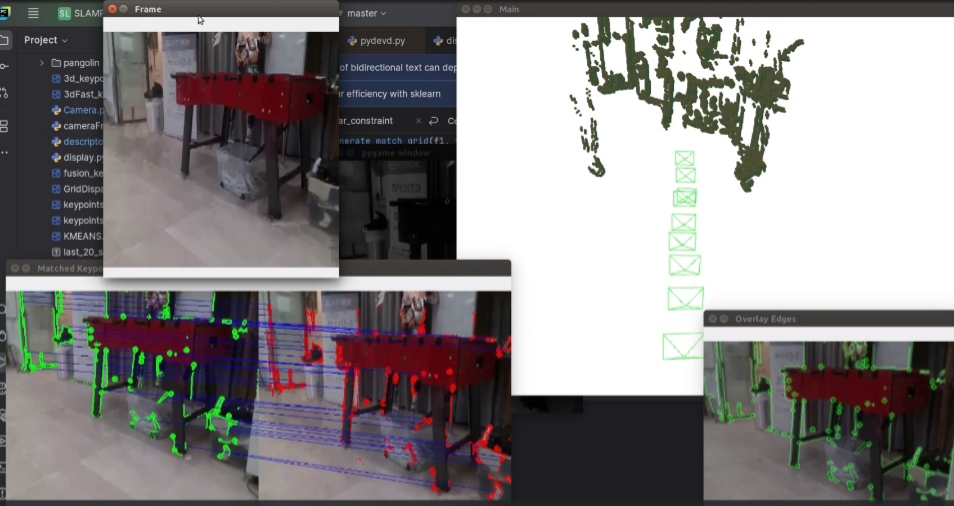}
		\label{fig:orbslam2_result}
	\end{minipage}
	\caption{(left) High-accuracy 3D Edge map from the start of the sequence and (right) the map remains precise throughout, with no drift or errors in later frames. This highlights the robustness and reliability of our mapping system.}
	\label{fig:lab_map}
\end{figure}

Subsequently, Figure~\ref{fig:full_map_tum} presents the complete map reconstructed after processing the entire sequence, emphasizing the system's robustness and ability to maintain high mapping accuracy over time. Lastly, Figure~\ref{fig:lab_map} displays the results within a laboratory environment, confirming the system's consistency and reliability in different settings.



\section{Conclusion}
\label{sec:Conclusion}
\noindent This work tackles real-time, accurate 3D mapping and navigation on resource-limited aerial platforms using monocular vision. We developed a lightweight, edge-aware SLAM system that combines sparse geometric features with dense edge and depth data, enabling dense semi-structured mapping without heavy global optimization. Our sensor fusion resolves scale ambiguity and ensures robustness indoors, avoiding neural networks that limit embedded deployment. Results show a 74.5\% improvement in RMSE for trajectory and pose estimation, with semi-dense maps more accurate and detailed. Additionally, our ORB-based edge map creation runs about 100 times faster than ORB-SLAM on a Ubuntu 16.04 laptop, facilitating real-time processing and deployment on small hardware like Raspberry Pi Zero. These findings demonstrate the effectiveness of combining classical geometric methods with deep learning-based depth and edge estimation for autonomous navigation. 
\section*{Acknowledgment}
J.D was supported by the Marine Hatter Foundation.

\bibliographystyle{unsrt}
\bibliography{mybib.bib}
\end{document}